\documentclass{article}
\usepackage{spconf,amsmath,graphicx}
\usepackage{amsfonts}
\usepackage{booktabs}
\usepackage[normalem]{ulem}
\usepackage{makecell}
\usepackage{multirow}
\usepackage{amssymb}
\usepackage{marvosym}
\useunder{\uline}{\ul}{}
\everymath{\displaystyle}

\title{LiSD: An Efficient Multi-Task Learning Framework for LiDAR Segmentation and Detection}
\name{Jiahua Xu$^{\star}$ \qquad Si Zuo$^{\dagger}$ \qquad Chenfeng Wei$^{\star}$\textsuperscript{\Letter} \qquad Wei Zhou$^{\ddagger}$\textsuperscript{\Letter}}
  
\address{$^{\star}$Wuxi Intelligent Control Research Institute, Hunan University, China\\
      $^{\dagger}$College of Mechanical and Vehicle Engineering, Hunan University, China\\
      $^{\ddagger}$School of Computer Science and Informatics, Cardiff University, United Kingdom}

\begin{document}
%
\maketitle
\begin{abstract}
With the rapid proliferation of autonomous driving, there has been a heightened focus on the research of lidar-based 3D semantic segmentation and object detection methodologies, aiming to ensure the safety of traffic participants. In recent decades, learning-based approaches have emerged, demonstrating remarkable performance gains in comparison to conventional algorithms. However, the segmentation and detection tasks have traditionally been examined in isolation to achieve the best precision. To this end, we propose an efficient multi-task learning framework named LiSD which can address both segmentation and detection tasks, aiming to optimize the overall performance. Our proposed LiSD is a voxel-based encoder-decoder framework that contains a hierarchical feature collaboration module and a holistic information aggregation module. Different integration methods are adopted to keep sparsity in segmentation while densifying features for query initialization in detection. Besides, cross-task information is utilized in an instance-aware refinement module to obtain more accurate predictions. Experimental results on the nuScenes dataset and Waymo Open Dataset demonstrate the effectiveness of our proposed model. It is worth noting that LiSD achieves the state-of-the-art performance of 83.3\% mIoU on the nuScenes segmentation benchmark for lidar-only methods.

\end{abstract}
\section{Introduction}
\label{sec:intro}

Semantic segmentation and 3D object detection tasks play pivotal roles in autonomous driving, serving as foundational components for establishing a comprehensive environmental perception system, which is crucial for mitigating labor costs and ensuring traffic safety. To facilitate advancements in this field, large-scale databases such as nuScenes \cite{caesar2020nuscenes} and Waymo Open Dataset (WOD) \cite{sun2020scalability} have been created, which are invaluable resources for guiding the refinement and evaluation of the perception algorithms designed to handle complicated road conditions. Based on the aforementioned databases, semantic segmentation and object detection tasks are conventionally researched independently to accomplish the best accuracy, for example, frameworks such as Cylinder3D are purposefully tailored for semantic segmentation, which can not yield remarkable performance compared to frameworks explicitly designed for object detection \cite{zhou2020cylinder3d}. Hence, there exists an urgent need to investigate unified frameworks for achieving optimal performance across both semantic segmentation and object detection tasks. Concurrently, the generation of segmentation and detection outcomes within a singular inference process proves time-saving, presenting an advantage over the separate execution of distinct tasks \cite{feng2021simple}.

\begin{figure}[t]
	\centerline{\includegraphics[width=8cm]{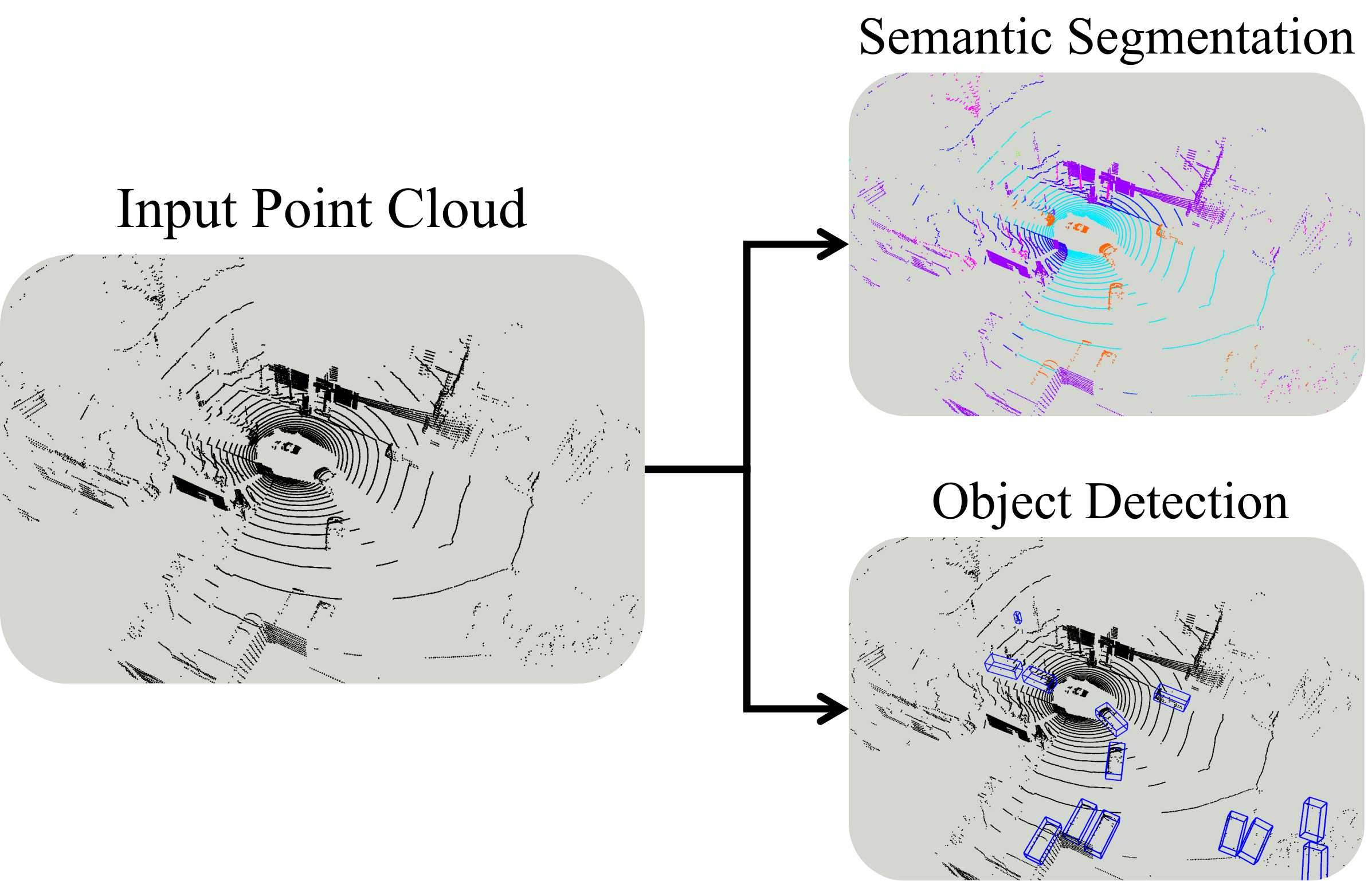}}
	\caption{The proposed LiSD model adopts point cloud data as its input and simultaneously produces semantic segmentation and object detection results.}
	\centering
	\label{fig:fig1}
\end{figure}

\begin{figure*}[t]
  \centerline{\includegraphics[width=18cm]{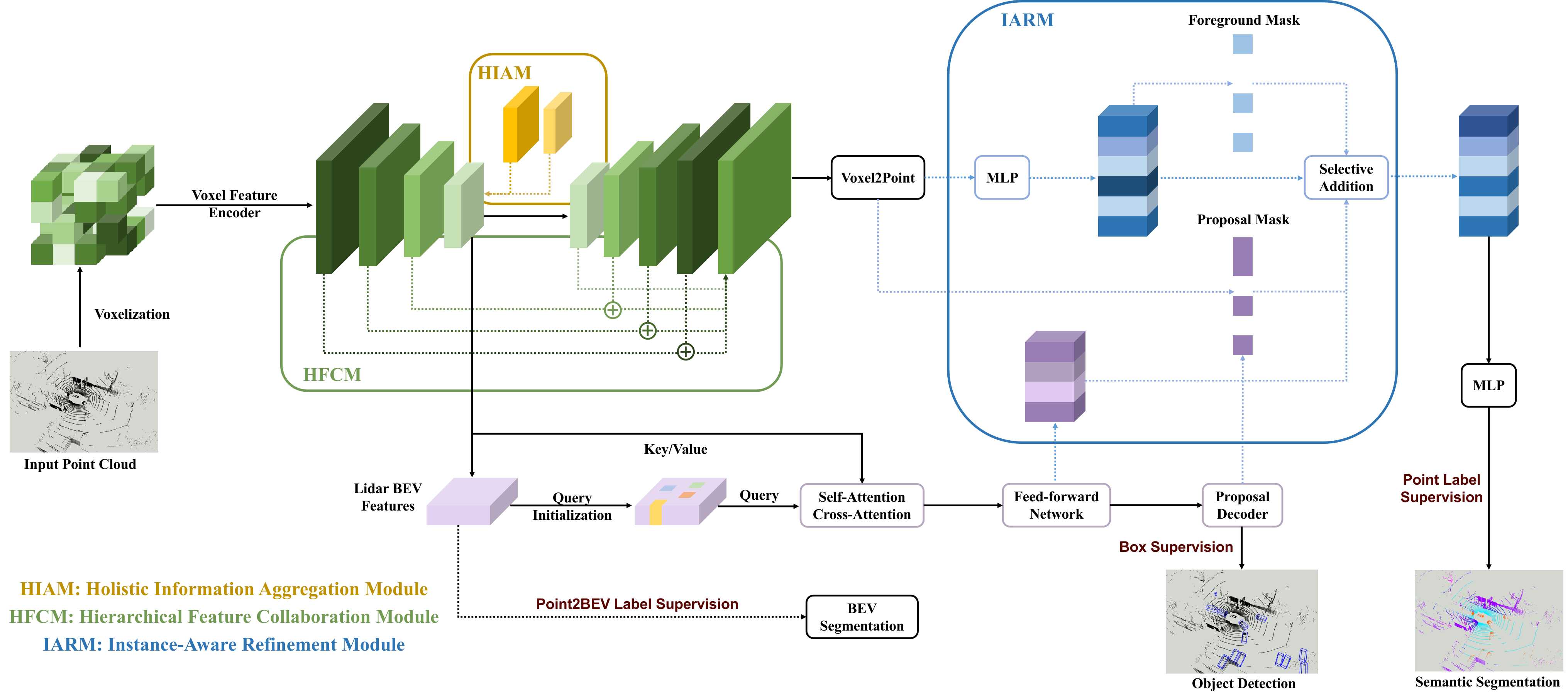}}
  \caption{The architecture of our proposed LiSD is outlined, where the point cloud serves as input for both segmentation and detection tasks. HIAM and HFCM are adopted in the voxel-based encoder-decoder to integrate holistic and hierarchical information for both tasks. Additionally, IARM is introduced to directly refine the point-level feature representation and indirectly exert influence on the box regression.}
  \centering
\label{fig:fig2}
\end{figure*}

Recently, deep learning frameworks have achieved considerable success in the domain of lidar perception, which can be roughly categorized into point-based, voxel-based, Range View (RV) based, Bird's Eye View (BEV) based, and hybrid methodologies. Voxel-based methods have emerged as the predominant paradigm in segmentation and detection tasks, owing to the development of sparse convolution techniques \cite{yan2018second}. Standard sparse convolution yields output points when there existing related points in the receptive field. In contrast, submanifold sparse convolution limits the output location as active only if the corresponding input location is active \cite{graham2017submanifold}. Submanifold convolution becomes indispensable in 3D networks to reduce memory consumption, which implies the receptive field is constrained when adopting submanifold convolution \cite{yan2018second}. VoxelNeXt \cite{chen2023voxelnext} incorporates additional down-sampling and sparse height compression to generate robust feature representation with sufficient receptive fields. Nevertheless, these operations will induce alterations in the density of the input sparse feature, consequently increasing the difficulty in the implementation of inverse sparse convolution. To address this problem, Ye \textit{et al.} propose the global context pooling (GCP) in the multi-task framework LidarMultiNet \cite{ye2023lidarmultinet}. GCP transforms the 3D sparse features into 2D BEV dense features to extract the global information. Moreover, a cross-space transformer module is adopted in LiDARFormer \cite{zhou2023lidarformer} to learn long-range information in the BEV feature. However, direct extraction of the global information from the conversion of 2D BEV dense features leads to an increase in memory consumption, given the storage of inactive points. 

Regarding the cross-task information interaction during multi-task learning, LidarMTL \cite{feng2021simple} adopts a conventional approach in which only low-level features are shared. In LidarMultiNet \cite{ye2023lidarmultinet}, a second-stage refinement module is introduced to enhance the first-stage semantic segmentation and produce the panoptic segmentation results. Zhou \textit{et al.} adopt a cross-task module \cite{zhou2023lidarformer} to transfer high-level features through cross-task attention mechanisms with high computational complexity.

In this paper, to reduce memory consumption and computational complexity while keeping precision, we propose an efficient multi-task learning framework, denoted as LiSD, for lidar semantic segmentation and object detection as shown in Fig. \ref{fig:fig1}. Instead of the direct placement of voxels from various scales onto the ground used in VoxelNeXt, we introduce a memory-friendly holistic information aggregation module, which interpolates the high-level features to the relevant active positions of low-level features. Through this methodology, sparsity is preserved with the acquisition of global information. Besides, the hierarchical feature collaboration is adopted in our LiSD to enhance the voxel feature representation. Moreover, in contrast to the aforementioned cross-task information interaction methodologies, we propose a straightforward yet effective instance-aware refinement module. This module is specifically designed to enhance the feature representation of foreground points through the incorporation of proposal features. LiSD is evaluated on two databases, namely nuScenes and WOD, demonstrating competitive performance for both segmentation and detection tasks. Notably, LiSD attains a leading segmentation performance of 83.3\% mIoU on nuScene, outperforming all the lidar-based methods currently ranked on the leaderboard.

The main contributions can be listed as follows:
\begin{itemize}
	\item We propose an efficient multi-task learning framework LiSD for lidar semantic segmentation and object detection.
	\item We introduce a memory-friendly holistic information aggregation module (HIAM) to integrate global information suitable for segmentation and detection tasks, and a hierarchical feature collaboration module (HFCM) to enhance the voxel feature representation.
	\item We present an instance-aware refinement module (IARM) to improve the foreground point feature representation with the assistance of object proposals.
        \item The proposed LiSD achieves competitive performance in segmentation and detection tasks on the nuScenes and WOD datasets.
\end{itemize}

\section{Method}
\label{sec:method}

In this section, we delineate the structure of our multi-task learning framework, LiSD, which seamlessly integrates three perception tasks, namely, semantic segmentation, object detection, and the auxiliary BEV segmentation through a single feed-forward pass, as illustrated in Fig. \ref{fig:fig2}. 

\subsection{Overview}
\label{ssec:overview}

Given the input point cloud $P=\left \{ p_i\mid p_i\in  \mathbb{R}^{3+c}  \right \} ^{N}_{i=1}$, the proposed LiSD yields semantic segmentation labels  $L=\left \{ l_i\mid l_i\in  \left ( 1\cdots K  \right )  \right \} ^{N}_{i=1}$ and object detection bounding boxes $B=\left \{ b_i\mid b_i\in  \mathbb{R}^{9}  \right \} ^M_{i=1}$,  where $N$ represents the number of points, $M$, $K$ denote the number of predicted boxes and semantic classes, respectively. Each point is endowed with $\left (3+c\right )$-dimensional features, e.g., 3D coordinate, intensity, elongation, timestamp, etc. The predicted boxes are characterized by the center coordinates, sizes, orientations, and velocities.

Firstly, the voxelized point cloud is fed into the Voxel Feature Encoder (VFE) for producing the sparse voxel feature representation $F_{v_j} = mean (p_i), i=1\cdots n$ through an average pooling layer, where $p_i$ denotes the feature representation of the $i$-th point within the voxel $v_j$, and $v_j$ contains a total of $n$ points. Then, the voxel-based encoder-decoder formed from 3D sparse convolution is adopted to generate voxel and BEV feature representation for segmentation and detection tasks. The encoder comprises four stages of sparse convolution blocks to downsample the spatial resolution, thereby acquiring high-level voxel features for the detection head. Conversely, the decoder is equipped with four symmetrical stages of sparse inverse convolution blocks to recover to the original voxel size for the segmentation head. The holistic information aggregation module and the hierarchical feature collaboration module are introduced in the encoder-decoder to enlarge the receptive field and enhance the feature representation. Ultimately, the segmentation and detection heads are employed to produce the semantic labels and object bounding boxes. An instance-aware refinement module is designed to integrate cross-task information to improve the accuracy of the prediction results.

\subsection{Holistic Information Aggregation Module}
\label{ssec:HIAM}

As inspired by VoxelNeXt \cite{chen2023voxelnext}, sufficient receptive fields are required to ensure correct predictions when dealing with sparse voxel features. Additional two stages of down-sampling are introduced in VoxelNeXt to generate features $\left \{ F_5, F_6\right \}$ with strides $\left \{ 16, 32\right \}$, and the multi-scale features of the original encoder with strides $\left \{ 1, 2, 4, 8\right \}$ are denoted as $\left \{ F_1, F_2, F_3, F_4\right \}$. The enhanced feature representation $F_d$ for the detection head is obtained as follows:
\begin{equation}\label{eq1}
\begin{aligned}
F_d & = F_4\cup F_5\cup F_6 \\
P_d & = P_4\cup P_5^{'} \cup P_6^{'}
\end{aligned}
\end{equation}
where $P_d$ represents the position of the enhanced voxel feature, and $P_4, P_5, P_6$ correspond to the positions of $F_4, F_5, F_6$, respectively. $P_5^{'}$ is aligned to $P_4$ by doubling the 3D coordinates $\left ( x_{p_{5}},  y_{p_{5}}, z_{p_{5}}  \right  )$ of the position $p_{5} \in P_{5}$, and the same as $P_6^{'}$. Nevertheless, the produced feature $F_d$ exhibits a significantly higher density compared to $F_4$. The different sparsity of $F_d$ and $F_4$ poses challenges in the implementation of inverse sparse convolution.

\begin{figure}[t]
	\centerline{\includegraphics[width=8.5cm]{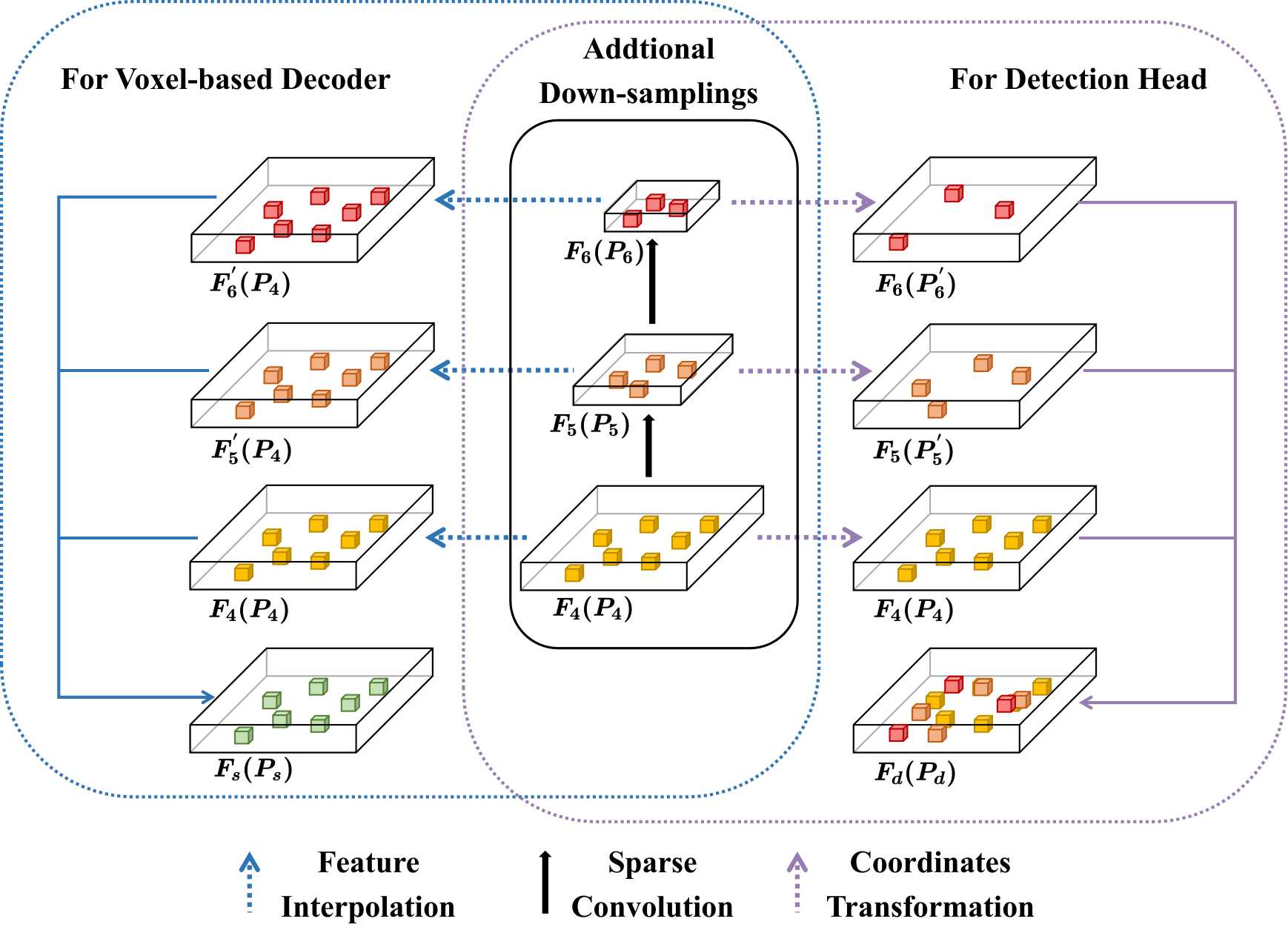}}
	\caption{The detailed structure of the proposed HIAM. Additional down-samplings are adopted to acquire holistic information. Subsequently, feature interpolation is employed to aggregate the information for the voxel-based decoder preceding the segmentation head, while coordinates transformation is utilized to integrate the information for the detection head.}
	\centering
	\label{fig:fig3}
\end{figure}

To solve this problem, we take another approach to integrate holistic information for the voxel-based decoder, as depicted in Fig. \ref{fig:fig3}. Specifically, the voxel features in $F_5^{'}$ corresponding to the position $P_4$ are interpolated with neighboring voxel features in $F_5$ to maintain sparsity, and this process is replicated for $F_6^{'}$. The enhanced feature representation $F_s$ for the voxel-based decoder is denoted as follows:
\begin{equation}\label{eq2}
\begin{aligned}
F_s & = F_4 + F_5^{'} + F_6^{'} \\
P_s & = P_4
\end{aligned}
\end{equation}
where $P_s$ denotes the position of the refined voxel feature, identical to $P_4$. Consequently, the implementation of inverse sparse convolution becomes straightforward. Note that $F_d$ is further projected to the BEV feature map $\overline{F}_d$ by putting all voxels onto the ground and summing up features in the same positions. Leveraging the HIAM, the proposed LiSD could significantly extend the receptive fields, which is a crucial aspect for improving semantic segmentation. Furthermore, the BEV feature map obtained through HIAM is further leveraged by object detection and BEV segmentation tasks.

\subsection{Hierarchical Feature Collaboration Module}
\label{ssec:HFCM}

As indicated in \cite{lin2017feature}, hierarchical features possess robust semantic information across various scales, which is beneficial for the semantic segmentation task. Apart from the 3D U-Net architecture, we design an additional hierarchical feature collaboration module to augment the voxel feature representation for the segmentation head. As previously noted, the multi-scale features of the encoder and decoder with strides $\left \{ 1, 2, 4, 8\right \}$ are represented as $\left \{ F_1, F_2, F_3, F_4\right \}$ and $\left \{ F_1^{'}, F_2^{'}, F_3^{'}, F_s\right \}$, respectively. The enhanced feature $\overline{F}_s$ is obtained via the top-down pathway and lateral connections as follows:
\begin{equation}\label{eq3}
\overline{F}_s = cat\left (  \theta_1  ( F_1+F_1^{'}), \theta_2  ( F_2+F_2^{'}), \theta_3 ( F_3+F_3^{'} ), \theta_s F_s \right ) 
\end{equation}
where $\theta_1$, $\theta_2$, $\theta_3$, $\theta_s$ represent the encoding and up-scaling functions applied to hierarchical features, and $cat$ denotes the concatenation operation across feature channels.

\subsection{Instance-Aware Refinement Module}
\label{ssec:IARM}

Given that the devised LiSD is a multi-task learning framework, there exists the potential for the integration of cross-task information between the segmentation and detection heads. As illustrated in Fig. \ref{fig:fig2}, voxel features undergo an initial conversion to point features for foreground probability estimation, which is guided by semantic segmentation labels during the training phase. The foreground mask ${m}_{fi}$ of the $i$-th point is denoted as:
\begin{equation}\label{eq4}
{m}_{fi}=\left\{\begin{matrix}\ \,1, \quad p_{fi}\ge \delta_f 
 \\
\ \,0,\quad p_{fi}< \delta_f 
\end{matrix}\right.
\end{equation}
where $p_{fi}$ represents the foreground probability of the $i$-th point, and $\delta_f$ indicates the probability threshold to distinguish between foreground and background points, e.g. $\delta_f=0.5$ in this paper. Simultaneously, the detection head predicts $M$ initial boxes, and the proposal mask ${m}_{bij}$ of the $i$-th point ($p_i$) to $j$-th box ($b_j$) is computed as:
\begin{equation}\label{eq5}
{m}_{bij}=\left\{\begin{matrix}1, \quad p_i \  inside \  b_j
 \\
\ \;0, \quad p_i \  outside \  b_j
\end{matrix}\right.
\end{equation}
Then, the refined point feature ${f}_{pi}^{'}$ of the $i$-th point is obtained as:
\begin{equation}\label{eq6}
f_{pi}^{'} = f_{pi}+\sum_{j=0}^{M} m_{fi}\cdot m_{bij} \cdot \mathrm{MLP} (f_{bj})
\end{equation}
where ${f}_{pi}$ represents the feature of the $i$-th point before instance-aware refinement, ${f}_{bj}$ represents the feature of the $j$-th box generated from the prediction head. MLP denotes the Multi-Layer Perceptron to adjust the feature dimension of ${f}_{bj}$ to align with that of ${f}_{pi}$. Benefiting from the IARM, the feature representation of the foreground points is enhanced with the incorporation of the proposal features, and constraints are applied to the box regression process concurrently.

\subsection{Joint Training}
\label{ssec:JT}

LiSD is trained in an end-to-end manner via a multi-task loss function. Specifically, for the semantic and BEV segmentation tasks, the optimization is guided by the cross-entropy loss and Lovasz loss \cite{berman2018lovasz}. For the detection task,  binary cross-entropy loss, $L_1$ loss, and IoU loss \cite{zhou2019iou} are employed to minimize the classification, regression, and IoU cost, respectively. Subsequently, the final loss $L$ is defined as a weighted sum of the task-specific losses:
\begin{equation}\label{eq6}
L = \sum_{i \in \left \{ seg, bev,det \right \} }\frac{1}{2\sigma_i^2 } L_i+\frac{1}{2}\mathrm{log}\sigma_i^2 
\end{equation}
where $\sigma_i$ denotes the noise parameter for task $i$ to compute task-dependent uncertainty \cite{kendall2018multi}. Hence, as the uncertainty increases for task $i$, the contribution of the task-specific loss $L_i$ to $L$ diminishes.

\begin{table*}[t]
\begin{center}
\caption{Segmentation results for each class on the \textit{test} split of the nuScenes dataset. The best and second-best performing results are marked in boldface and underlined.}
\label{table1}
\setlength{\tabcolsep}{2.5mm}{
\scalebox{0.82}{
\begin{tabular}{@{}c|c|cccccccccccccccc@{}}
\toprule
Model         & mIoU          & \rotatebox{80}{barrier}       & \rotatebox{80}{bicycle}       & \rotatebox{80}{bus}           & \rotatebox{80}{car}           & \rotatebox{80}{construction}  & \rotatebox{80}{motorcycle}    & \rotatebox{80}{pedestrian}    & \rotatebox{80}{traffic cone}  & \rotatebox{80}{trailer}       & \rotatebox{80}{truck}         & \rotatebox{80}{driveable} & \rotatebox{80}{other flat}    & \rotatebox{80}{sidewalk}      & \rotatebox{80}{terrain}       & \rotatebox{80}{manmade}       & \rotatebox{80}{vegetation}    \\ \midrule
PolarNet \cite{zhang2020polarnet}      & 69.8          & 80.1          & 19.9          & 78.6          & 84.1          & 53.2                 & 47.9          & 70.5          & 66.9          & 70.0          & 56.7          & 96.7              & 68.7          & 77.7          & 72.0          & 88.5          & 85.4          \\
PolarStream \cite{chen2021polarstream}      & 73.4          & 71.4          & 27.8          & 78.1          & 82.0          & 61.3                 & 77.8          & 75.1          & 72.4          & 79.6          & 63.7          & 96.0              & 66.5          & 76.9          & 73.0          & 88.5          & 84.8          \\
SPVNAS \cite{tang2020searching}     & 77.4          & 80.0          & 30.0          & {\ul 91.9}    & 90.8   & 64.7                 & 79.0          & 75.6          & 70.9          & 81.0          & 74.6 & 97.4              & 69.2 & 80.0    & 76.1          & 89.3          & 87.1\\
Cylinder3D++ \cite{zhou2020cylinder3d} & 77.9          & 82.8          & 33.9          & 84.3          & 89.4          & 69.6                 & 79.4          & 77.3          & 73.4          & 84.6          & 69.4          & 97.7              & {\ul 70.2}          & 80.3          & 75.5          & 90.4          & 87.6          \\
AF2S3Net \cite{cheng20212}      & 78.3          & 78.9          & 52.2          & 89.9          & 84.2          & {\ul 77.4}           & 74.3          & 77.3          & 72.0          & 83.9          & 73.8          & 97.1              & 66.5          & 77.5          & 74.0          & 87.7          & 86.8          \\
SPVCNN++ \cite{tang2020searching}     & 81.1          & 86.4          & 43.1          & {\ul 91.9}    & {\ul 92.2}    & 75.9                 & 75.7          & 83.4          & 77.3          & 86.8          & \textbf{77.4} & 97.7              & \textbf{71.2} & {\ul 81.1}    & 77.2          & 91.7          & 89.0          \\
LidarMultiNet \cite{ye2023lidarmultinet} & 81.4          & 80.4          & 48.4          & \textbf{94.3} & 90.0          & 71.5                 & {\ul 87.2}    & {\ul 85.2}    & {\ul 80.4}    & 86.9          & 74.8          & {\ul 97.8}        & 67.3          & 80.7          & 76.5          & 92.1          & 89.6          \\
LidarFormer \cite{zhou2023lidarformer}   & 81.5          & {\ul 84.4}    & 40.8          & 84.7          & \textbf{92.6} & 72.7                 & \textbf{91.0} & 84.9          & \textbf{81.7} & \textbf{88.6} & 73.8          & \textbf{97.9}     & 69.3          & \textbf{81.4} & \textbf{77.4}    & {\ul 92.4}    & 89.6          \\
UDeerPep      & {\ul 81.8}    & \textbf{85.5} & {\ul 55.5}    & 90.5          & 91.6          & 72.2                 & 85.6          & 81.4          & 76.3          & {\ul 87.3}    & 74.0          & 97.7              & {\ul 70.2}          & {\ul 81.1}    & \textbf{77.4}    & \textbf{92.7} & \textbf{90.2} \\ \midrule
Proposed LiSD          & \textbf{83.3} & 82.1          & \textbf{67.1} & 89.8          & {\ul 92.2}    & \textbf{80.5}        & 86.9          & \textbf{87.4} & 79.3          & 86.6          & {\ul 76.1}    & 97.5              & 67.2          & 80.5          & 77.0          & 92.3          & {\ul 89.7}          \\ \bottomrule
\end{tabular}}}
\end{center}
\end{table*}

\section{Experiments}
\label{sec:exp}

In this section, the databases used in our experiment are introduced at first. Then, we conduct the performance comparison of LiSD and other methods on these databases. Finally, the effectiveness of the specially designed modules is verified via the ablation study.

\subsection{Datasets}
\label{ssec:dataset}

Two large-scale autonomous driving databases equipped with point-wise semantic labels and 3D object bounding box annotations are utilized in our experiment, namely nuScenes dataset \cite{caesar2020nuscenes} and WOD dataset \cite{sun2020scalability}.

\textbf{NuScenes dataset \cite{caesar2020nuscenes}}: This database includes 1000 scenarios, each lasting 20 seconds and captured using a 32-beam lidar sensor at a sampling rate of 20Hz. Keyframes within each scenario are annotated with a 2Hz sampling rate. For the semantic segmentation task, point-wise semantic labels are provided for 16 categories, including 10 foreground classes and 6 background classes, and mean Intersection over Union (mIoU) is employed as the evaluation metric. For the object detection task, bounding box annotations are provided for the same 10 foreground categories as those in the segmentation task, and the evaluation metrics include mean Average Precision (mAP) and NuScenes Detection Score (NDS).

\textbf{WOD dataset \cite{sun2020scalability}}: This dataset contains 2000 scenarios captured by a 64-beam lidar sensor at a sampling rate of 10Hz. Similar to nuScenes, each scenario spans 20 seconds, with detection annotations available for all frames, while point-wise semantic labels are only provided for selected keyframes. For the semantic segmentation task, semantic labels are provided for 23 classes, and the evaluation metric employed is mIoU as well. For the object detection task, bounding box annotations are provided for 3 categories, e.g., vehicles, pedestrians, and cyclists. Average Precision Weighted by Heading (APH) is used as the evaluation metric for detection, and the ground truth objects are categorized as LEVEL$\_$1 (L1) and LEVEL$\_$2 (L2) samples based on the detection difficulty. 
mAPH L1 is calculated by considering samples labeled as L1, while mAPH L2 is computed by incorporating both L1 and L2 samples.

\subsection{Experiment Setup}
\label{ssec:setup}

The AdamW optimizer, coupled with a one-cycle scheduler, is adopted to train the proposed LiSD for 65 epochs, with a max learning rate of 3e-3 and a weight decay of 0.01. The voxel size of cloud range $\left [ -54.0,54.0 \right ]m \times \left [ -54.0,54.0 \right ]m \times \left [ -5.0,4.6 \right ]m$ is set as $\left [0.075,0.075,0.2 \right ]$ for nuScenes, and voxel size of cloud range $\left [ -75.2,75.2 \right ]m \times \left [ -75.2,75.2 \right ]m \times \left [ -2,4 \right ]m$ is set as $\left [0.1,0.1,0.15 \right ]$ for WOD. Standard data augmentation techniques, including flipping, scaling, rotation, translation, and ground-truth sampling \cite{yan2018second} with fade strategy \cite{wang2021pointaugmenting} are utilized during training. In our experiment, 6 Nvidia A30 GPUs with a batch size of 18 are employed for NuScenes, while a batch size of 12 is configured for WOD.

\begin{table}[ht]
\caption{Comparison results on the \textit{val} split of the nuScenes dataset. Reported by \cite{zhou2023lidarformer}. The best and second-best performing results are marked in boldface and underlined.}
\label{table2}
\setlength{\tabcolsep}{4.5mm}{
\scalebox{0.82}{
\begin{tabular}{@{}c|c|c|cc@{}}
\toprule
                              & Model        & mIoU          & mAP           & NDS           \\ \midrule
\multirow{3}{*}{Segmentation} & PolarNet \cite{zhang2020polarnet}        & 71.0          & -             & -             \\
                              & Cylinder3D \cite{zhou2020cylinder3d}    & 76.1          & -             & -             \\
                              & RPVNet \cite{xu2021rpvnet}        & 77.6          & -             & -             \\ \midrule
\multirow{3}{*}{Detection}    & CenterPoint \cite{yin2021center}   & -             & 57.4          & 65.2          \\
                              & VoxelNeXt \cite{chen2023voxelnext}   & -             & 60.5          & 66.6          \\
                              & TransFusion-L \cite{bai2022transfusion} & -             & 60.0          & 66.8          \\ \midrule
\multirow{3}{*}{Multi-task}   & LidarMultiNet \cite{ye2023lidarmultinet} & 82.0         &  63.8          & {\ul69.5}    \\
                              & LidarFormer \cite{zhou2023lidarformer}          & {\ul82.7} & \textbf{66.6}    & \textbf{70.8}    \\ 
                              & Proposed LiSD          & \textbf{83.0} &  {\ul 65.0}   & {\ul69.5}    \\ \bottomrule
\end{tabular}}}
\end{table}

\subsection{Experiment Results}
\label{ssec:results}

Segmentation and detection results for both the nuScenes and WOD datasets are presented to substantiate the effectiveness of LiSD.

\textbf{NuScenes dataset}: Performance comparisons of LiSD and other SOTA methodologies are listed in Table \ref{table1}, and we can observe that LiSD achieves top segmentation performance of 83.3\% mIoU on the \textit{test} split of nuScenes. The mIoU of LiSD is 1.5\% higher than that of UDeerPep, establishing its superiority over the best-performing lidar-based methods currently positioned atop the leaderboard. In terms of multi-task models, a second stage is regarded as unnecessary in LiSD when compared to LidarMultiNet \cite{ye2023lidarmultinet}. Besides, LiSD demonstrates competitive performance when contrasted with the complicated cross-space and cross-task transformers featured in LidarFormer \cite{zhou2023lidarformer}. Furthermore, the detection and segmentation performance on the \textit{val} split is illustrated in Table \ref{table2}. LiSD outperforms models tailored for segmentation tasks, including PolarNet \cite{zhang2020polarnet}, Cylinder3D \cite{zhou2020cylinder3d}, and RPVNet \cite{xu2021rpvnet}, in terms of mIoU. Concurrently, LiSD surpasses models specifically designed for detection tasks, such as CenterPoint \cite{yin2021center}, VoxelNeXt \cite{chen2023voxelnext}, and TransFusion-L \cite{bai2022transfusion}, by achieving higher mAP.

\begin{table}[ht]
\caption{Comparison results on the \textit{val} split of the WOD dataset. Reported by \cite{zhou2023lidarformer}. The best and second-best performing results are marked in boldface and underlined.}
\label{table3}
\setlength{\tabcolsep}{5.5mm}{
\scalebox{0.82}{
\begin{tabular}{@{}c|c|c|c@{}}
\toprule
                              & Model        & mIoU          & L2mAPH              \\ \midrule
\multirow{2}{*}{Segmentation} & PolarNet \cite{zhang2020polarnet}        &61.6         & -             \\
                              & Cylinder3D \cite{zhou2020cylinder3d}    & 66.6          & -         \\ \midrule
\multirow{3}{*}{Detection}    & CenterPoint++ \cite{yin2021center}   & -             & 71.6          \\
                              & CenterFormer \cite{zhou2022centerformer}   & -             & 73.7          \\
                              & MPPNet \cite{chen2022mppnet} & -             & 74.9               \\ \midrule
\multirow{3}{*}{Multi-task}   & LidarMultiNet \cite{ye2023lidarmultinet} & 71.9         &  75.2      \\
                              & LidarFormer \cite{zhou2023lidarformer}          & {\ul72.2} & \textbf{76.2}     \\ 
                              & Proposed LiSD          & \textbf{ 72.6} &  {\ul 76.1}    \\ \bottomrule
\end{tabular}}}
\end{table}

\textbf{WOD dataset}: Table \ref{table3} illustrates the performance comparison of semantic segmentation and object detection on the \textit{val} split of the WOD dataset. The segmentation results of PolarNet \cite{zhang2020polarnet} and Cylinder3D \cite{zhou2020cylinder3d} are reproduced by \cite{zhou2023lidarformer}. As observed in Table \ref{table3}, the proposed LiSD attains a mIoU of 72.6\% for the segmentation task and an L2mAPH of 76.1\% for the detection task, outperforming the single-task model. Meanwhile, LiSD exhibits competitive performance in comparison with multi-task models, such as LidarMultiNet \cite{ye2023lidarmultinet} and LidarFormer \cite{zhou2023lidarformer}.

The experimental results of LiSD on both datasets illustrate that multi-task learning facilitates the interaction of information across tasks, thereby contributing to the high performance for both tasks.

\subsection{Ablation study}
\label{ssec:ablation}

The ablation study, depicted in Table \ref{table4}, systematically validates the effectiveness of each key component within our proposed LiSD. The baseline segmentation model employed in our experiment is based on the standard 3D sparse U-Net architecture, and the baseline detection model is Transfusion-L \cite{bai2022transfusion}. The performance improvements brought by specially designed modules, including HIAM, HFCM, and IARM are further verified on the \textit{val} split of the nuScenes dataset.

\begin{table}[ht]
\caption{Ablation study for improvement of mIoU and mAP on the \textit{val} split of the nuScenes dataset.}
\label{table4}
\setlength{\tabcolsep}{1.5mm}{
\scalebox{0.82}{
\begin{tabular}{@{}ccccccc|cc@{}}
\toprule
Baseline & HIAM & HFCM & \begin{tabular}[c]{@{}c@{}}Multi\\ Task\end{tabular} & \begin{tabular}[c]{@{}c@{}}BEV\\ Loss\end{tabular} & IARM & TTA & mIoU & mAP  \\ \midrule
\checkmark        &      &      &                                                      &                                                    &      &     & 77.1 & 60.0 \\
\checkmark        & \checkmark    &      &                                                      &                                                    &      &     & 77.9 & 61.7 \\
\checkmark        & \checkmark    & \checkmark    &                                                      &                                                    &      &     & 79.0 & 62.1 \\
\checkmark        & \checkmark    & \checkmark    & \checkmark                                                    &                                                    &      &     & 80.4 & 62.8 \\
\checkmark        & \checkmark    & \checkmark    & \checkmark                                                    & \checkmark                                                  &      &     & 81.1 & 64.1 \\
\checkmark        & \checkmark    & \checkmark    & \checkmark                                                    & \checkmark                                                  & \checkmark    &     & 81.9 & 64.5 \\
\checkmark        & \checkmark    & \checkmark    & \checkmark                                                    & \checkmark                                                  & \checkmark    & \checkmark   & \textbf{83.0} & \textbf{65.0} \\ \bottomrule
\end{tabular}}}
\end{table}

\begin{figure}[ht]
	\centerline{\includegraphics[width=8.3cm]{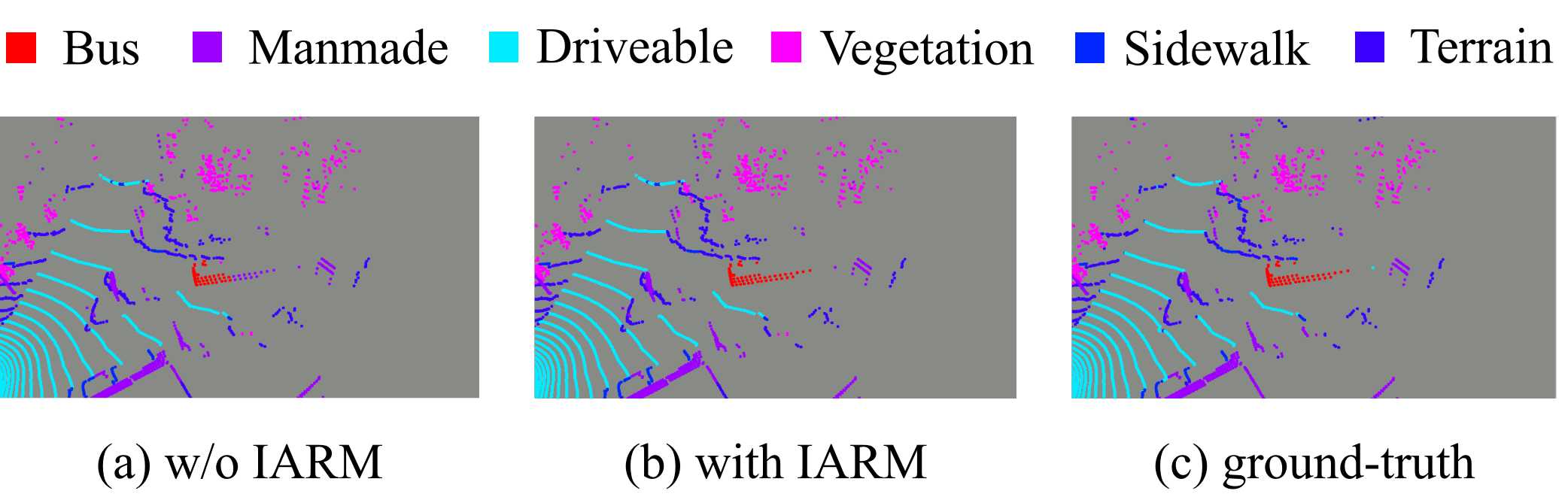}}
	\caption{Semantic segmentation results on the nuScenes dataset. (a) Predicted results of LiSD without IARM, (b) Predicted results of LiSD with IARM, (c) Ground-truth segmentation labels.}
	\centering
	\label{fig:fig4}
\end{figure}

As we can observe from Table \ref{table4}, the baseline model achieves a mIoU of 77.1\% and a mAP of 60.0\% on the validation set. Building upon the baseline model, the incorporation of the HIAM aimed at enlarging the receptive field yields an improvement of 0.8\% in mIoU and 1.7\% in mAP. Incorporating the HFCM leads to an increase of 1.1\% in mIoU and 0.4\% in mAP. The combination of segmentation and detection tasks further improves mIoU by 1.4\% and mAP by 0.7\%. Besides, through the utilization of BEV segmentation loss coupled with uncertainty weighting, the mIoU and mAP increase by 0.7\% and 1.3\%. The IARM, designed for the collaboration of cross-task information, brings 0.8\% mIoU and 0.4\% mAP improvement. The visualization results are depicted in Fig. \ref{fig:fig4}.  Forming our optimal model on the validation set, the implementation of Test Time Augmentation (TTA) improves the mIoU and mAP to 83.0\% and 65.0\%, respectively.

\section{Conclusion}
\label{sec:conclusion}

In this paper, we propose an efficient multi-task learning framework named LiSD for lidar segmentation and detection, which are predominantly addressed separately in previous works. Comprehensive experimental results verified the effectiveness of LiSD's design, including key components such as HIAM, HFCM, and IARM. Moreover, LiSD achieves a higher mIoU compared to the top-performing lidar-based method currently positioned on the leaderboard of nuScenes lidar segmentation task. We hope that our proposed LiSD can serve as an inspiration for future endeavors in the development of multi-modal multi-task learning frameworks.

\vfill\pagebreak



\small
\bibliographystyle{IEEEbib}
\bibliography{refs}

\begin{thebibliography}{10}

\bibitem{caesar2020nuscenes}
Holger Caesar, Varun Bankiti, Alex~H Lang, Sourabh Vora, Venice~Erin Liong, Qiang Xu, Anush Krishnan, Yu~Pan, Giancarlo Baldan, and Oscar Beijbom,
\newblock ``nuscenes: A multimodal dataset for autonomous driving,''
\newblock in {\em Proceedings of the IEEE/CVF conference on computer vision and pattern recognition}, 2020, pp. 11621--11631.

\bibitem{sun2020scalability}
Pei Sun, Henrik Kretzschmar, Xerxes Dotiwalla, Aurelien Chouard, Vijaysai Patnaik, Paul Tsui, James Guo, Yin Zhou, Yuning Chai, Benjamin Caine, et~al.,
\newblock ``Scalability in perception for autonomous driving: Waymo open dataset,''
\newblock in {\em Proceedings of the IEEE/CVF conference on computer vision and pattern recognition}, 2020, pp. 2446--2454.

\bibitem{zhou2020cylinder3d}
Xinge Zhu, Hui Zhou, Tai Wang, Fangzhou Hong, Yuexin Ma, Wei Li, Hongsheng Li, and Dahua Lin,
\newblock ``Cylindrical and asymmetrical 3d convolution networks for lidar segmentation,''
\newblock in {\em Proceedings of the IEEE/CVF conference on computer vision and pattern recognition}, 2021, pp. 9939--9948.

\bibitem{feng2021simple}
Di~Feng, Yiyang Zhou, Chenfeng Xu, Masayoshi Tomizuka, and Wei Zhan,
\newblock ``A simple and efficient multi-task network for 3d object detection and road understanding,''
\newblock in {\em 2021 IEEE/RSJ International Conference on Intelligent Robots and Systems (IROS)}. IEEE, 2021, pp. 7067--7074.

\bibitem{yan2018second}
Yan Yan, Yuxing Mao, and Bo~Li,
\newblock ``Second: Sparsely embedded convolutional detection,''
\newblock {\em Sensors}, vol. 18, no. 10, pp. 3337, 2018.

\bibitem{graham2017submanifold}
Benjamin Graham and Laurens Van~der Maaten,
\newblock ``Submanifold sparse convolutional networks,''
\newblock {\em arXiv preprint arXiv:1706.01307}, 2017.

\bibitem{chen2023voxelnext}
Yukang Chen, Jianhui Liu, Xiangyu Zhang, Xiaojuan Qi, and Jiaya Jia,
\newblock ``Voxelnext: Fully sparse voxelnet for 3d object detection and tracking,''
\newblock in {\em Proceedings of the IEEE/CVF Conference on Computer Vision and Pattern Recognition}, 2023, pp. 21674--21683.

\bibitem{ye2023lidarmultinet}
Dongqiangzi Ye, Zixiang Zhou, Weijia Chen, Yufei Xie, Yu~Wang, Panqu Wang, and Hassan Foroosh,
\newblock ``Lidarmultinet: Towards a unified multi-task network for lidar perception,''
\newblock in {\em Proceedings of the AAAI Conference on Artificial Intelligence}, 2023, vol.~37, pp. 3231--3240.

\bibitem{zhou2023lidarformer}
Zixiang Zhou, Dongqiangzi Ye, Weijia Chen, Yufei Xie, Yu~Wang, Panqu Wang, and Hassan Foroosh,
\newblock ``Lidarformer: A unified transformer-based multi-task network for lidar perception,''
\newblock {\em arXiv preprint arXiv:2303.12194}, 2023.

\bibitem{lin2017feature}
Tsung-Yi Lin, Piotr Doll{\'a}r, Ross Girshick, Kaiming He, Bharath Hariharan, and Serge Belongie,
\newblock ``Feature pyramid networks for object detection,''
\newblock in {\em Proceedings of the IEEE conference on computer vision and pattern recognition}, 2017, pp. 2117--2125.

\bibitem{berman2018lovasz}
Maxim Berman, Amal~Rannen Triki, and Matthew~B Blaschko,
\newblock ``The lov{\'a}sz-softmax loss: A tractable surrogate for the optimization of the intersection-over-union measure in neural networks,''
\newblock in {\em Proceedings of the IEEE conference on computer vision and pattern recognition}, 2018, pp. 4413--4421.

\bibitem{zhou2019iou}
Dingfu Zhou, Jin Fang, Xibin Song, Chenye Guan, Junbo Yin, Yuchao Dai, and Ruigang Yang,
\newblock ``Iou loss for 2d/3d object detection,''
\newblock in {\em 2019 international conference on 3D vision (3DV)}. IEEE, 2019, pp. 85--94.

\bibitem{kendall2018multi}
Alex Kendall, Yarin Gal, and Roberto Cipolla,
\newblock ``Multi-task learning using uncertainty to weigh losses for scene geometry and semantics,''
\newblock in {\em Proceedings of the IEEE conference on computer vision and pattern recognition}, 2018, pp. 7482--7491.

\bibitem{zhang2020polarnet}
Yang Zhang, Zixiang Zhou, Philip David, Xiangyu Yue, Zerong Xi, Boqing Gong, and Hassan Foroosh,
\newblock ``Polarnet: An improved grid representation for online lidar point clouds semantic segmentation,''
\newblock in {\em Proceedings of the IEEE/CVF Conference on Computer Vision and Pattern Recognition}, 2020, pp. 9601--9610.

\bibitem{chen2021polarstream}
Qi~Chen, Sourabh Vora, and Oscar Beijbom,
\newblock ``Polarstream: Streaming object detection and segmentation with polar pillars,''
\newblock {\em Advances in Neural Information Processing Systems}, vol. 34, pp. 26871--26883, 2021.

\bibitem{tang2020searching}
Haotian Tang, Zhijian Liu, Shengyu Zhao, Yujun Lin, Ji~Lin, Hanrui Wang, and Song Han,
\newblock ``Searching efficient 3d architectures with sparse point-voxel convolution,''
\newblock in {\em European conference on computer vision}. Springer, 2020, pp. 685--702.

\bibitem{cheng20212}
Ran Cheng, Ryan Razani, Ehsan Taghavi, Enxu Li, and Bingbing Liu,
\newblock ``2-s3net: Attentive feature fusion with adaptive feature selection for sparse semantic segmentation network,''
\newblock in {\em Proceedings of the IEEE/CVF conference on computer vision and pattern recognition}, 2021, pp. 12547--12556.

\bibitem{wang2021pointaugmenting}
Chunwei Wang, Chao Ma, Ming Zhu, and Xiaokang Yang,
\newblock ``Pointaugmenting: Cross-modal augmentation for 3d object detection,''
\newblock in {\em Proceedings of the IEEE/CVF Conference on Computer Vision and Pattern Recognition}, 2021, pp. 11794--11803.

\bibitem{xu2021rpvnet}
Jianyun Xu, Ruixiang Zhang, Jian Dou, Yushi Zhu, Jie Sun, and Shiliang Pu,
\newblock ``Rpvnet: A deep and efficient range-point-voxel fusion network for lidar point cloud segmentation,''
\newblock in {\em Proceedings of the IEEE/CVF International Conference on Computer Vision}, 2021, pp. 16024--16033.

\bibitem{yin2021center}
Tianwei Yin, Xingyi Zhou, and Philipp Krahenbuhl,
\newblock ``Center-based 3d object detection and tracking,''
\newblock in {\em Proceedings of the IEEE/CVF conference on computer vision and pattern recognition}, 2021, pp. 11784--11793.

\bibitem{bai2022transfusion}
Xuyang Bai, Zeyu Hu, Xinge Zhu, Qingqiu Huang, Yilun Chen, Hongbo Fu, and Chiew-Lan Tai,
\newblock ``Transfusion: Robust lidar-camera fusion for 3d object detection with transformers,''
\newblock in {\em Proceedings of the IEEE/CVF conference on computer vision and pattern recognition}, 2022, pp. 1090--1099.

\bibitem{zhou2022centerformer}
Zixiang Zhou, Xiangchen Zhao, Yu~Wang, Panqu Wang, and Hassan Foroosh,
\newblock ``Centerformer: Center-based transformer for 3d object detection,''
\newblock in {\em European Conference on Computer Vision}. Springer, 2022, pp. 496--513.

\bibitem{chen2022mppnet}
Xuesong Chen, Shaoshuai Shi, Benjin Zhu, Ka~Chun Cheung, Hang Xu, and Hongsheng Li,
\newblock ``Mppnet: Multi-frame feature intertwining with proxy points for 3d temporal object detection,''
\newblock in {\em European Conference on Computer Vision}. Springer, 2022, pp. 680--697.

\end{thebibliography}
\footnote{© 2024 IEEE. Personal use of this material is permitted. Permission from IEEE must be obtained for all other uses, in any current or future media, including reprinting/republishing this material for advertising or promotional purposes, creating new collective works, for resale or redistribution to servers or lists, or reuse of any copyrighted component of this work in other works.}
\end{document}